\documentclass{article}

\usepackage[preprint]{neurips_2024}

\usepackage[utf8]{inputenc} % allow utf-8 input
\usepackage[T1]{fontenc}    % use 8-bit T1 fonts
\usepackage{hyperref}       % hyperlinks
\usepackage{url}            % simple URL typesetting
\usepackage{booktabs}       % professional-quality tables
\usepackage{amsfonts}       % blackboard math symbols
\usepackage{nicefrac}       % compact symbols for 1/2, etc.
\usepackage{microtype}      % microtypography
\usepackage{xcolor}         % colors

\usepackage{graphicx} % Required for inserting images
\usepackage{subcaption}
\usepackage{amsmath}
\usepackage{amssymb}
\usepackage{float}

\title{Probe-based Fine-tuning for reducing Toxicity}
\author{Jan Wehner, Mario Fritz}
\date{October 2025}

\author{%
  Jan Wehner \qquad Mario Fritz \\
  CISPA Helmholtz Center for Information Security\\
}

\begin{document}

\maketitle

\begin{abstract}
Probes trained on model activations can detect undesirable behaviors like deception or biases that are difficult to identify from outputs alone. This makes them useful detectors to identify misbehavior. Furthermore, they are also valuable training signals, since they not only reward outputs, but also good internal processes for arriving at that output. However, training against interpretability tools raises a fundamental concern: when a monitor becomes a training target, it may cease to be reliable (Goodhart's Law). We propose two methods for training against probes based on Supervised Fine-tuning and Direct Preference Optimization. We conduct an initial exploration of these methods in a testbed for reducing toxicity and evaluate the amount by which probe accuracy drops when training against them. To retain the accuracy of probe-detectors after training, we attempt (1) to train against an ensemble of probes, (2) retain held-out probes that aren't used for training, and (3) retrain new probes after training.

First, probe-based preference optimization unexpectedly preserves probe detectability better than classifier-based methods, suggesting the preference learning objective incentivizes maintaining rather than obfuscating relevant representations. Second, probe diversity provides minimal practical benefit—simply retraining probes after optimization recovers high detection accuracy. Our findings suggest probe-based training can be viable for certain alignment methods, though probe ensembles are largely unnecessary when retraining is feasible.\footnote{Code available at \url{https://github.com/janweh/probe_based_training}}
\end{abstract}

\section{Introduction}
Language models can exhibit undesirable behaviors that are difficult to detect from outputs alone, such as deception, internal reward hacking, or subtle forms of toxicity. Recent work has shown that linear probes trained on model activations can detect these behaviors with high accuracy, even when they are not apparent in the model's outputs. This has motivated interest in using probes as a monitor to detect when the model is misbehaving.

However, probing the internal activations of a model could also serve as a valuable training signal. By detecting wrong behavior that is not detectable from the outputs alone, activation-based scores can serve as a new modality for training objectives. As a motivating example, we can take honesty. From the outputs alone, it is very difficult to detect whether a model is answering truthfully or not. However, lie-detector probes can identify dishonesty with very high accuracy \cite{goldowskydill2025detectingstrategicdeceptionusing}. We can thus use the scores given by the lie-detector probes as a penalty when training the model in order to incentivize it to become more honest.

However, training against interpretability tools introduces a fundamental challenge: when a monitor becomes a training target, it may cease to be a reliable monitor. This phenomenon, an instance of Goodhart's Law, has been documented in recent work on monitoring reasoning models and lie detection \cite{goldowskydill2025detectingstrategicdeceptionusing}. Models can learn to evade specific monitors while continuing the undesired behavior in ways the probe cannot detect. Indeed, the approach of training against your interpretability techniques has been coined ``The Most Forbidden Technique'' \cite{mowshowitz2025forbidden}. This raises a critical question: can we train against probes while retaining the ability to detect the problematic behavior?

We investigate whether using diverse ensembles of probes can address this challenge. Our approach trains models against one set of probes while evaluating whether independently trained "held-out" probes retain their detection accuracy. We focus on toxicity reduction as a tractable testbed, training the Gemma-3-1B model using both supervised fine-tuning (SFT) and direct preference optimization (DPO) with probe-based regularization.
Our key findings are twofold. First, probe-based DPO unexpectedly preserves probe accuracy better than classifier-based DPO, suggesting that optimizing against internal representations may strengthen rather than destroy the signal. Second, while ensemble methods provide modest improvements, we find that simply retraining probes after optimization often recovers similar detection accuracy, limiting the practical need for probe diversity in this setting.

These results contribute to ongoing discussions about the viability of interpretability-based training methods. While our work identifies some scenarios where probe-based training shows promise, concurrent work by Gupta and Jenner demonstrates that architecturally robust probes (attention-based and max-pooling) may offer a more scalable solution to probe evasion. Our findings help delineate when probe diversity matters and when other approaches may be more effective.

\section{Related Work}

\textbf{Training Against Internal Representations.} 
Recent work has explored using model internals as training signals. \cite{papadatos2024linearprobepenaltiesreduce} demonstrated that penalizing probe predictions during training can reduce sycophancy in language models, providing early evidence that probe-based regularization can modify undesirable behaviors. \cite{chen2025personavectorsmonitoringcontrolling} showed that steering vectors derived from model activations can both monitor and control character traits, establishing that internal representations contain manipulable signals for behavioral traits like toxicity.

\textbf{Probe Evasion and Goodhart's Law.}
A central concern with probe-based training is whether models learn to evade detection rather than eliminate the problematic behavior—an instance of Goodhart's Law where a measure ceases to be useful once it becomes a target. Two concurrent works directly address this question. \cite{cundy2025preferencelearningliedetectors} show that preference learning with lie detectors can induce either genuine honesty or sophisticated evasion depending on the training setup, demonstrating that probe-based training outcomes are highly sensitive to implementation details. Such evasion behavior has previously been shown for training against Chain-of-Thought monitors \cite{baker2025monitoringreasoningmodelsmisbehavior}. \cite{gupta2025rlobfuscationlanguagemodelslearn} systematically study whether language models learn to evade latent-space monitors during RL training, finding that standard linear probes are vulnerable to evasion but that architecturally robust probes (attention-based and max-pooling) resist such gaming. \cite{bailey2025obfuscatedactivationsbypassllm} further demonstrate that models can learn to obfuscate their activations to bypass latent-space defenses, reinforcing concerns about probe reliability under optimization pressure.

Our work provides complementary empirical evidence on probe evasion in the specific context of toxicity reduction, comparing probe-based versus classifier-based preference optimization and testing whether probe ensembles mitigate evasion. While \cite{gupta2025rlobfuscationlanguagemodelslearn} identify architectural robustness as a promising solution, our results on probe diversity and retraining offer an alternative perspective on when standard linear probes remain viable monitoring tools.

\section{Methods}
\subsection{Training Toxicity Probes}
We trained linear probes to detect toxicity in the model's internal representations. Each probe consists of a logistic regression classifier applied to activations from a specific layer of the language model. The probe training pipeline follows a standardized procedure to ensure consistent evaluation across different experimental conditions.

\subsubsection{Activation Collection}

For each training sample, we collect activations during the generation phase. Given a toxic prompt from the Civil Comments dataset \cite{duchene2023benchmarktoxiccommentclassification}, we generate a response using greedy decoding (no sampling) with a maximum of 64 new tokens. We extract the hidden states from all 34 layers of the model at each generation step, then compute the mean activation across all generated tokens (excluding the initial prompt) to obtain a single activation vector per layer per sample. This mean-pooling strategy provides a representation that captures the overall toxicity signal throughout the generation process.

Each generated response is scored using the RoBERTa toxicity classifier~\cite{Detoxify}, which provides a continuous toxicity score between 0 and 1. Samples with scores $\geq 0.5$ are labeled as toxic (positive class), while those below this threshold are labeled as non-toxic (negative class).

\subsubsection{Training Procedure}

Each probe is trained on 500 samples from a distinct subset of the toxic comments distribution. We use a two-stage pipeline: first, activations are standardized to have zero mean and unit variance; second, a logistic regression classifier with L2 regularization is trained on the standardized activations.

When training with multiple layers, we train a separate probe for each layer and select the best-performing layer based on validation set AUC. The selected layer (layer 20 for Gemma-3-1B) is then used consistently across all experiments. This layer selection ensures that the probe operates on representations where toxicity information is most linearly separable.

After training, each probe is calibrated on a held-out validation set of 50 samples to find the decision boundary that yields the highest accuracy.

The resulting probes are diverse, as is shown in their low cosine similarity in Appendix \ref{app:diversity}.

\subsection{Probe-based Direct Preference Optimization}
\label{sec:dpo_method}

\begin{figure}
    \centering
    \includegraphics[width=\linewidth]{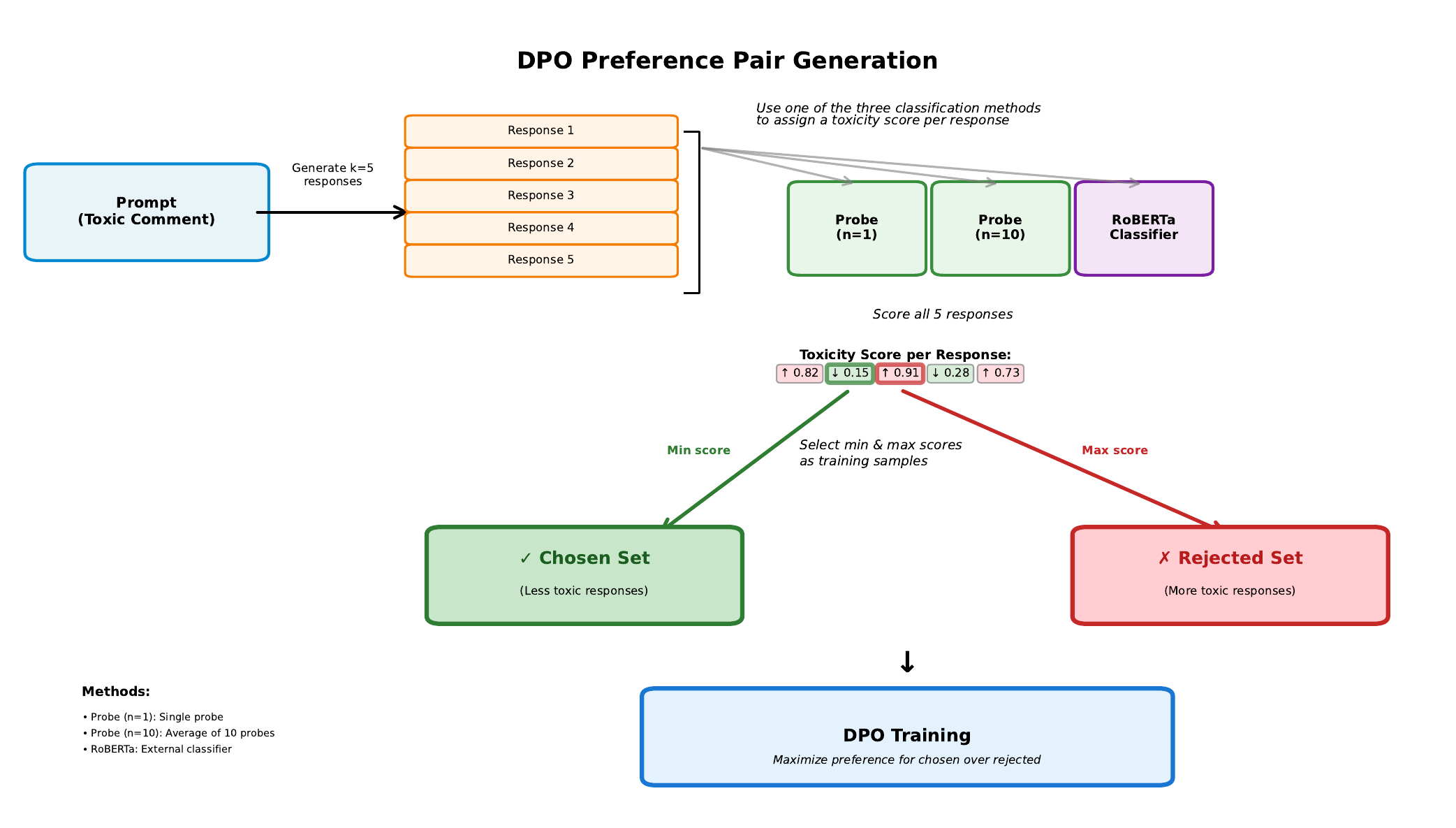}
    \caption{Probe-based DPO method}
    \label{fig:dpo_scheme}
\end{figure}

Direct Preference Optimization (DPO)~\cite{rafailov2024directpreferenceoptimizationlanguage} trains language models to prefer certain outputs over others by directly optimizing on preference pairs, without requiring an explicit reward model. For each prompt in a dataset, we need both a preferred and rejected answer. While this is normally done with a text-based classifier (our baseline), we instead judge responses based on their probe-score. See Figure \ref{fig:dpo_scheme} for a visualization of the method.

\subsubsection{Preference Pair Generation}

For each training prompt, we generate $k=5$ candidate responses using the base model with sampling enabled (temperature = 1.0). Each candidate is then scored for toxicity to determine which responses should be preferred (chosen) versus rejected. We compare two scoring approaches:

\textbf{Probe-based scoring:} We extract the mean activation vector from the selected layer (layer 20) across all generated tokens for each candidate response. The trained probe predicts a toxicity probability for each activation vector. The candidate with the lowest probe probability is labeled as the "chosen" response, while the candidate with the highest probe probability is labeled as "rejected."

\textbf{Classifier-based scoring:} As a baseline, we use the RoBERTa toxicity classifier to directly score each generated text. The candidate with the lowest classifier score is labeled as "chosen," and the one with the highest score is labeled as "rejected."

This process generates a dataset of preference pairs $\mathcal{D} = \{(x_i, y_i^{\text{chosen}}, y_i^{\text{rejected}})\}$ where $x_i$ is the prompt, $y_i^{\text{chosen}}$ is the less toxic response, and $y_i^{\text{rejected}}$ is the more toxic response. We generate pairs from 10,000 training prompts, resulting in a dataset split into 70\% training, 20\% validation, and 10\% test sets.

\subsubsection{DPO Training}

We fine-tune the base model using the standard DPO objective~\cite{rafailov2024directpreferenceoptimizationlanguage}:

\begin{equation}
\mathcal{L}_{\text{DPO}}(\pi_\theta; \pi_{\text{ref}}) = -\mathbb{E}_{(x,y^+,y^-) \sim \mathcal{D}} \left[ \log \sigma \left( \beta \log \frac{\pi_\theta(y^+ | x)}{\pi_{\text{ref}}(y^+ | x)} - \beta \log \frac{\pi_\theta(y^- | x)}{\pi_{\text{ref}}(y^- | x)} \right) \right]
\end{equation}

where $\pi_\theta$ is the policy being optimized, $\pi_{\text{ref}}$ is the frozen reference model (the base model), $y^+$ and $y^-$ denote the chosen and rejected responses respectively, $\beta=0.1$ is the KL penalty coefficient, and $\sigma$ is the sigmoid function. We use  LoRA~\cite{hu2022lora} for parameter-efficient fine-tuning. %with rank $r=8$, $\alpha=32$, and dropout of 0.05, targeting the attention projection and feed-forward layers. Training uses a batch size of 4 per device with 4 gradient accumulation steps, a learning rate of $1\times10^{-5}$, and runs for 1 epoch. We apply label smoothing of 0.05 to prevent overconfident predictions.

\subsubsection{Experimental Conditions}

We evaluate three DPO conditions:

\begin{itemize}
    \item \textbf{DPO-Classifier:} Standard DPO using preference pairs as judged by the RoBERTa toxicity classifier. This serves as our baseline and does not involve probes.
    \item \textbf{Single DPO-Probe:} DPO using preference pairs as judged by a single trained probe.
    \item \textbf{10 DPO-Probes:} DPO using preference pairs as judged by combining predictions from 10 independently trained probes (averaged probabilities).
\end{itemize}

We don't expect probe-based DPO to outperform output-based DPO for training against toxicity, since toxicity can be well detected from the outputs. Our experiments using toxicity should be taken more as a proof of concept that probe-based training can successfully reduce toxicity. Further research should evaluate the success of this training scheme in preventing deception and dishonesty.

\subsection{Supervised Fine-Tuning with Probe-Based Regularization}
\label{sec:sft_method}

We introduce a probe-based regularization approach for supervised fine-tuning (SFT) that penalizes the model for generating representations that the probe classifies as toxic. This method directly incorporates toxicity detection into the training objective, steering the model away from toxic generation patterns at the representation level.

This could be useful if one wants to train on a dataset that contains toxic content (eg an unfiltered scrape from social media) and wants to learn the information in it without making the model more toxic.

\subsubsection{Training Objective}

The standard language modeling objective for SFT is the negative log-likelihood:
\begin{equation}
\mathcal{L}_{\text{LM}} = -\mathbb{E}_{x \sim \mathcal{D}} \left[ \log p_\theta(x) \right]
\end{equation}
where $x$ is a training sequence and $\theta$ represents the model parameters.

We augment this objective with a probe-based penalty term. During training, we extract the hidden state activations $h \in \mathbb{R}^{d}$ from the selected layer (layer 20) and compute the mean-pooled activation across all tokens in each training batch. The probe then predicts a toxicity probability $p_{\text{probe}}(h)$ for these activations. Our combined training objective becomes:

\begin{equation}
\mathcal{L}_{\text{total}} = \mathcal{L}_{\text{LM}} + \lambda \cdot p_{\text{probe}}(h)
\end{equation}

where $\lambda$ is a hyperparameter controlling the strength of the probe-based regularization. Importantly, the probe parameters are frozen during SFT—only the language model parameters $\theta$ are updated. This ensures gradients flow through the probe to update the model's representations without modifying the probe itself.

We fine-tune the model using LoRA~\cite{hu2022lora} for 1 epoch on 10,000 samples from the Civil Comments training set.

\subsubsection{Multi-Probe Aggregation}
When training with multiple probes, there are different ways of aggregating their scores into a single score. We experiment with this in Appendix \ref{app:strategies}. Our experiments find that it's best to use one probe at a time and cycle them sequentially through training batches. This leverages the diversity of multiple independently trained probes while maintaining deterministic training dynamics. In the same experiment, we also evaluate how much priority to give to the probe-loss and find $\lambda=1$ to be a good value.

\section{Experiment}

\subsection{Dataset and Task}

We conduct our experiments using the Civil Comments dataset \cite{duchene2023benchmarktoxiccommentclassification}, a large-scale toxicity detection corpus containing approximately 2 million public comments from online discussions. We focus exclusively on comments labeled as toxic (toxicity score $\geq 0.5$) to create a challenging training environment where the model must learn to respond non-toxically to inherently toxic prompts. This design choice tests whether our probe-based methods can effectively steer model behavior when confronted with adversarial inputs that might naturally elicit toxic responses.

Each comment is formatted into a conversational prompt using the template: ``Respond to the following user comment: [comment text]''. This framing simulates a realistic scenario where language models must handle toxic user inputs while maintaining appropriate responses.

We use the Gemma-3-1B model \cite{gemmateam2025gemma3technicalreport}, which has not had alignment training applied. This is important for the experiments, as the instruction-tuned model barely shows toxic responses.

\subsection{Evaluation Metrics}

We track multiple metrics to assess both toxicity reduction and potential side effects:

\textbf{Toxicity Rate}: The fraction of model generations classified as toxic (score $\geq 0.5$) by the RoBERTa toxicity classifier on 100 held-out toxic prompts. Lower values indicate more successful detoxification.

\textbf{Probe AUC - Training Probes}: Area under the ROC curve for the probes used during training, evaluated on post-training activations. This measures whether the model's internal representations remain detectable as toxic according to the original training signal.

\textbf{Probe AUC - Held-Out Probes}: AUC for 2 independently trained probes (trained on separate data subsets) never exposed during model training. High AUC here indicates the model hasn't simply learned to fool the training probes but has genuinely altered its representations.

\textbf{Probe AUC - Retrained Probes}: AUC for 2 probes retrained from scratch on post-training activations from held-out data. This tests whether toxicity remains linearly detectable in the model's representation space after training. Low AUC suggests the model has fundamentally reorganized how it represents toxic content.

\textbf{Perplexity on the training distribution}: Measured on 1,000 held-out toxic comments from Civil Comments. This assesses whether probe-based training impedes the model's ability to fit the training distribution.

\textbf{Perplexity on Wiki}: Measured on 1,000 samples from WikiText-2. This assesses whether probe-based training degrades general language modeling capabilities on out-of-distribution clean text.

\section{Results}
This section shows the experimental results. It should be noted that results are noisy and were only run on one seed, thus limiting the strength of conclusions we can draw.

\subsection{DPO}
\begin{figure}[!b]
    \centering
        \begin{subfigure}[t]{0.3\textwidth}
        \includegraphics[width=\textwidth]{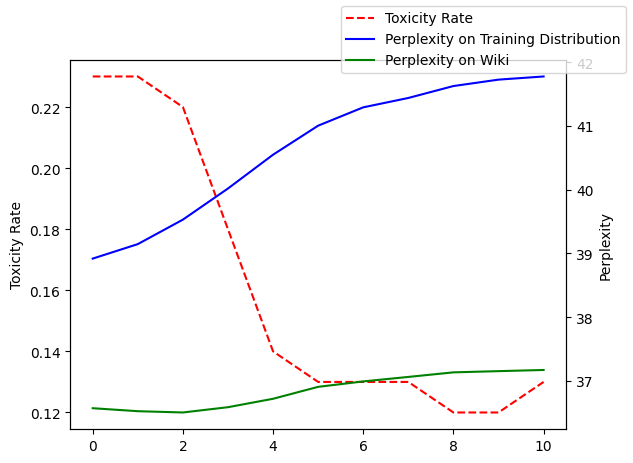}
        \caption{Toxicity rate, perplexity on training distribution, and perplexity on the Wiki-dataset during DPO training using a \textbf{text-based toxicity classifier}.}
    \end{subfigure}
    \hfill
    \begin{subfigure}[t]{0.3\textwidth}
        \includegraphics[width=\textwidth]{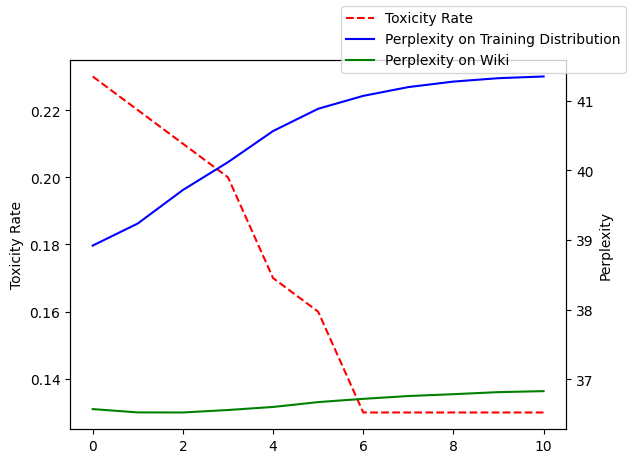}
        \caption{Toxicity rate, perplexity on training distribution, and perplexity on the Wiki-dataset during DPO training using \textbf{one probe-based classifier}.}
    \end{subfigure}
    \hfill
    \begin{subfigure}[t]{0.3\textwidth}
        \includegraphics[width=\textwidth]{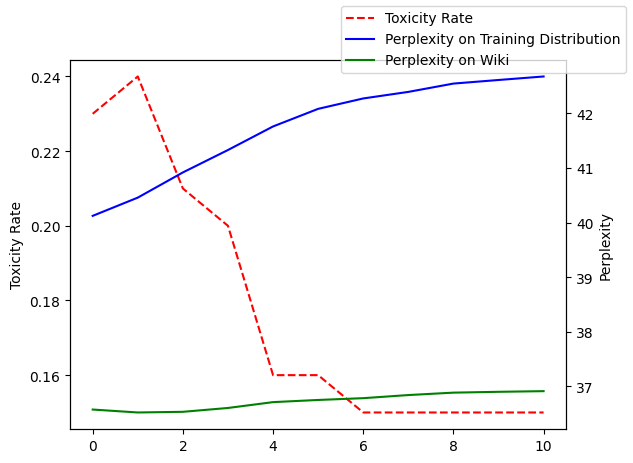}
        \caption{Toxicity rate, perplexity on training distribution, and perplexity on the Wiki-dataset during DPO training using \textbf{10 probe-based classifiers}.}
    \end{subfigure}

    \vspace{0.5cm} % spacing between rows
    
    \begin{subfigure}[t]{0.3\textwidth}
        \includegraphics[width=\textwidth]{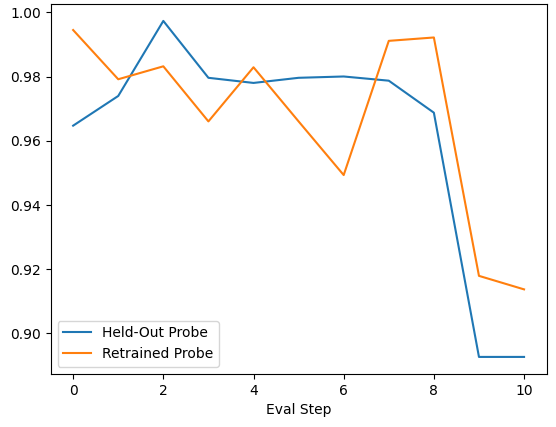}
        \caption{Area Under Curve (AUC) of held-out and newly trained probes during DPO training using a \textbf{text-based toxicity classifier}.}
    \end{subfigure}
    \hfill
    \begin{subfigure}[t]{0.3\textwidth}
        \includegraphics[width=\textwidth]{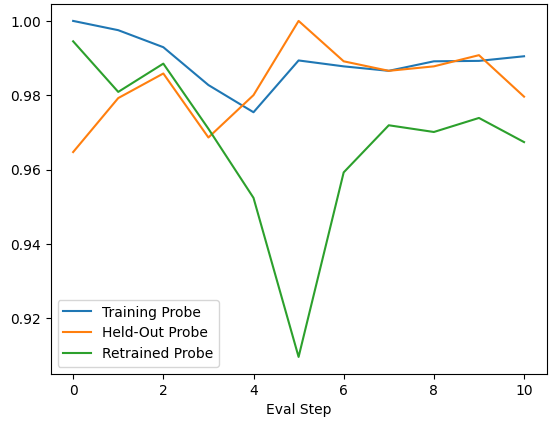}
        \caption{Area Under Curve (AUC) of the probe used for training, held-out, and newly trained probes during DPO training using \textbf{one probe-based classifier}.}
    \end{subfigure}
    \hfill
    \begin{subfigure}[t]{0.3\textwidth}
        \includegraphics[width=\textwidth]{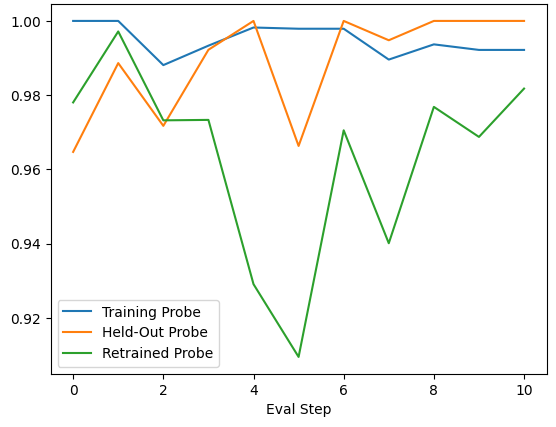}
        \caption{Area Under Curve (AUC) of probes used for training, held-out, and newly trained probes during DPO training using \textbf{10 probe-based classifiers}.}
    \end{subfigure}
    \caption{Changes in probe accuracy, toxicity rate, and model capability when training with DPO using a text-based classifier, one probe-based classifier, or 10 probe-based classifiers.}
    \label{fig:DPO_figures}
\end{figure}

We present results from three DPO conditions: classifier-based (baseline), single probe-based, and 10 probe-based. Table \ref{tab:DPO_table} shows final metrics, while Figure \ref{fig:DPO_figures} displays training dynamics.

\begin{table}[!b]
    \centering
    \begin{tabular}{c p{1.2cm} p{1.1cm} p{1.5cm} p{1.5cm} p{2cm} p{2cm}}
           
& Toxicity Rate&AUC train probes&AUC held-out probes&AUC retrained probes&Perplexity Training Distribution&Perplexity Wiki\\
\hline
Classifier&0.11&-&0.866&0.770&41.78&37.18\\
1 probe&0.14&0.977&0.938&0.926&41.35&37.03\\
10 probes&0.15&0.987&0.994&0.992&41.42&36.91\
    \end{tabular}
    \caption{Final results of DPO training.}
    \label{tab:DPO_table}
\end{table}

\textbf{Toxicity Reduction:} All methods substantially reduce toxicity from baseline (0.23). Classifier-based DPO achieves 0.11, while probe-based methods reach 0.14-0.15. Probe-based DPO produces 27-36\% more toxic outputs than the classifier baseline, but still delivers 35-40\% improvement over baseline.
\textbf{Probe Detectability:} Training against probes preserves detectability better than training against classifiers—a surprising reversal.
Classifier-based DPO degrades probe accuracy substantially: held-out probes drop to 0.866 AUC, retrained probes to 0.770. In contrast, single-probe DPO maintains 0.938 (held-out) and 0.926 (retrained), while 10-probe DPO achieves 0.994 and 0.992, respectively. Critically, training probes themselves retain 0.977-0.987 AUC, indicating the model doesn't evade its training signal through Goodhart-style obfuscation.
We hypothesize probe-based DPO may preserve detectability because the preference learning objective requires discriminating between chosen and rejected responses based on probe scores. This creates incentives to maintain—rather than destroy—the representational features probes detect. Classifier-based DPO lacks this incentive since the model never sees the representations underlying classifier judgments.

Using 10 probes versus 1 improves held-out probe AUC from 0.938 to 0.994 (90\% error reduction). Whether this justifies a 10× computational cost depends on application requirements.

General capabilities remain intact. WikiText-2 perplexity hovers around 37 for all methods (Table \ref{tab:DPO_table}). Training distribution perplexity increases from ~39 to ~41 across all methods—an expected DPO side-effect where the KL penalty pushes the policy away from the reference distribution.

\subsection{SFT}

\begin{figure}[htbp]
    \centering
        \begin{subfigure}[t]{0.3\textwidth}
        \includegraphics[width=\textwidth]{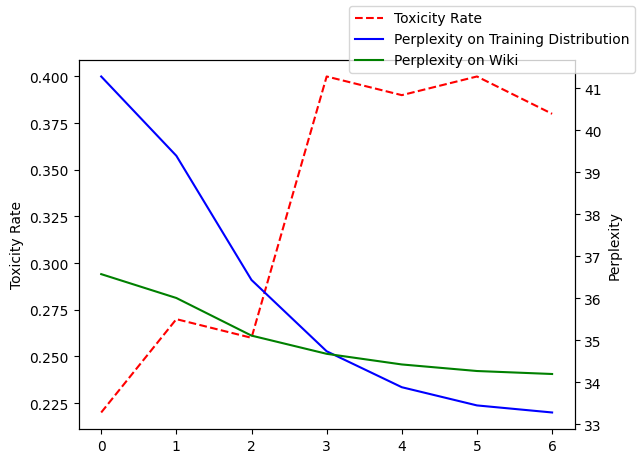}
        \caption{Toxicity rate, perplexity on training distribution, and perplexity on the Wiki-dataset during SFT training using a \textbf{text-based toxicity classifier}.}
    \end{subfigure}
    \hfill
    \begin{subfigure}[t]{0.3\textwidth}
        \includegraphics[width=\textwidth]{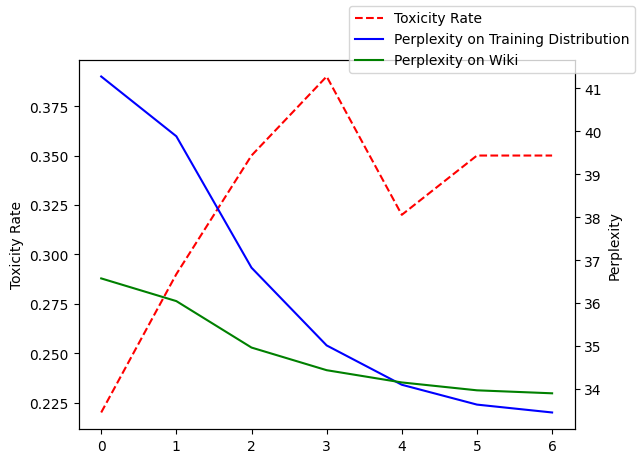}
        \caption{Toxicity rate, perplexity on training distribution, and perplexity on the Wiki-dataset during SFT training using \textbf{one probe-based classifier.}}
    \end{subfigure}
    \hfill
    \begin{subfigure}[t]{0.3\textwidth}
        \includegraphics[width=\textwidth]{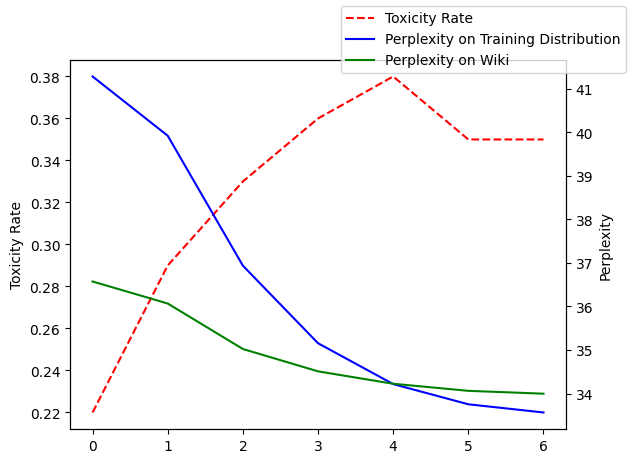}
        \caption{Toxicity rate, perplexity on training distribution, and perplexity on the Wiki-dataset during SFT training using \textbf{10 probe-based classifiers}.}
    \end{subfigure}

    \vspace{0.5cm} % spacing between rows
    
    \begin{subfigure}[t]{0.3\textwidth}
        \includegraphics[width=\textwidth]{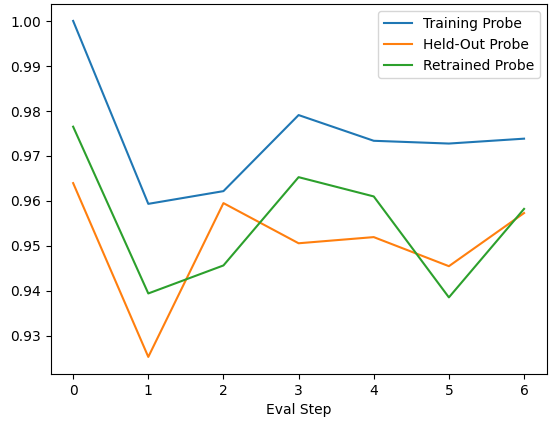}
        \caption{Area Under Curve (AUC) of held-out and newly trained probes during SFT training using a \textbf{text-based toxicity classifier}.}
    \end{subfigure}
    \hfill
    \begin{subfigure}[t]{0.3\textwidth}
        \includegraphics[width=\textwidth]{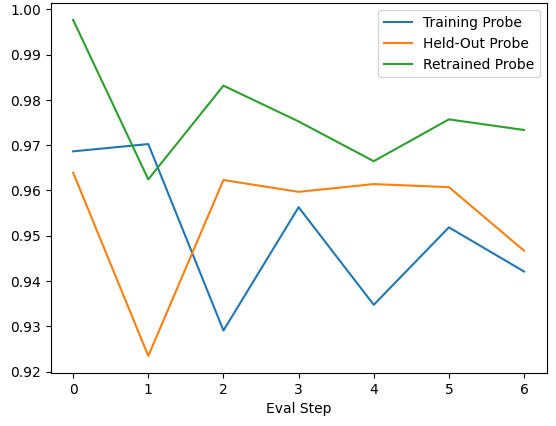}
        \caption{Area Under Curve (AUC) of the probe used for training, held-out and newly trained probes during SFT training using \textbf{one probe-based classifier}.}
    \end{subfigure}
    \hfill
    \begin{subfigure}[t]{0.3\textwidth}
        \includegraphics[width=\textwidth]{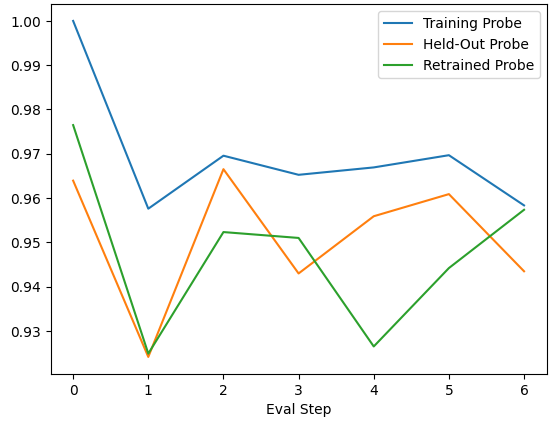}
        \caption{Area Under Curve (AUC) of probes used for training, held-out, and newly trained probes during SFT training using \textbf{10 probe-based classifiers}.}
    \end{subfigure}
    \caption{Changes in probe accuracy, toxicity rat,e and model capability when training with SFT using a text-based classifier, one probe-based classifier, or 10 probe-based classifiers.}
    \label{fig:SFT_figures}
\end{figure}

We present results from three SFT conditions: baseline (standard SFT without probe regularization), single probe-based regularization, and 10 probe-based regularization. Table \ref{tab:SFT_table} shows final metrics, while Figure \ref{fig:SFT_figures} displays training dynamics.

Standard SFT on toxic prompts increases toxicity from 0.22 (base model) to 0.38, as the model overfits to the toxic training distribution. Probe-based regularization partially mitigates this increase, with both single-probe and 10-probe training reaching 0.35—still 59\% more toxic than the base model, but representing an 8\% relative reduction compared to standard SFT.
These modest improvements contrast sharply with DPO results (0.11-0.15 toxicity). This aligns with SFT's intended use case from Section \ref{sec:sft_method}: preventing increases in toxicity when training on unfiltered data, rather than actively detoxifying the model.

Unlike DPO, SFT does not show a clear advantage for probe-based methods in preserving detectability. All methods maintain high probe AUC across evaluation types.
Baseline SFT achieves 0.974 (training probes), 0.957 (held-out), and 0.958 (retrained). Probe-based methods show modest degradation in training probe AUC (0.942-0.958), suggesting the model learns to partially evade the specific probes used during training. However, held-out probes (0.943-0.947) and especially retrained probes (0.957-0.973) maintain relatively high accuracy. This contrasts with DPO, where probe-based methods dramatically outperformed the baseline in preserving detectability. We speculate this difference arises because SFT's gradient-based regularization allows direct evasion of probe features, while DPO's preference learning objective incentivizes maintaining those features for discrimination.
Importantly, retraining probes after optimization recovers detection accuracy to near-baseline levels (0.957-0.973). This indicates toxicity remains linearly separable in representation space after training, just along slightly different directions. This result, consistent with DPO findings, suggests probe ensembles are unnecessary when retraining is feasible.
Training against probes reduces their AUC, but not by much. We believe this is because fine-tuning pushes the probe scores to become smaller, but the relative ordering of toxic/non-toxic examples stays the same. Thus, refitting the threshold to a lower value restores much of the accuracy.

Probe regularization doesn't impede language modeling. All methods achieve similar reductions in training perplexity (top panels of Figure 3), indicating the model successfully learns from the training data despite the probe penalty. WikiText-2 perplexity remains around 34 across all conditions (Table 2), confirming general capabilities are unaffected.
Multi-probe ensembles provide no advantage, with both single and 10-probe methods achieving identical toxicity rates (0.35) and comparable probe AUCs.

\begin{table}
    \centering
    \begin{tabular}{c p{1.2cm} p{1.1cm} p{1.5cm} p{1.5cm} p{2cm} p{2cm}}
           
& Toxicity Rate&AUC train probes&AUC held-out probes&AUC retrained probes&Perplexity Training Distribution&Perplexity Wiki\\
\hline
Classifier& 0.38&0.974&0.957&0.958&33.279&34.195\\
1 probe& 0.35&0.942&0.947&0.973&33.448&33.895\\
10 probes& 0.35&0.958&0.943&0.957&33.566&33.997\\
    \end{tabular}
    \caption{Final results of SFT training.}
    \label{tab:SFT_table}
\end{table}

\section{Conclusion}
We investigated whether training language models against internal representations inevitably leads to probe evasion—a central concern for interpretability-based alignment methods. Using toxicity reduction as a testbed, we compared probe-based and classifier-based training across supervised fine-tuning and direct preference optimization.

Our main finding is that the training method determines probe viability. Probe-based DPO preserves detectability substantially better than classifier-based DPO, maintaining high held-out probe accuracy while the baseline shows severe degradation. We hypothesize this stems from DPO's preference learning objective, which incentivizes maintaining—rather than obfuscating—the representational features that distinguish chosen from rejected responses. This suggests that optimizing against internal representations can strengthen rather than destroy the monitoring signal when the training objective aligns with probe detection.

In contrast, probe-based SFT shows no such advantage. While probe regularization modestly reduces toxicity increase during training on toxic data, it neither outperforms baseline methods nor prevents partial probe evasion. This method-specific difference suggests probe-based training is most viable for preference-based objectives rather than gradient-based regularization.

Our second finding limits the practical need for probe diversity. Multi-probe ensembles offer marginal improvements over single probes in both training methods. More importantly, probes retrained after optimization consistently recover high detection accuracy, indicating the targeted behavior remains linearly separable in representation space despite training pressure. When probe retraining is feasible, held-out ensembles provide little additional value.

These results should be interpreted alongside concurrent work showing that architectural robustness (attention-based and max-pooling probes) offers a more scalable solution to probe evasion \cite{gupta2025rlobfuscationlanguagemodelslearn}, and that probe-based preference learning outcomes are highly sensitive to implementation details \cite{cundy2025preferencelearningliedetectors}. Our contribution delineates when standard linear probes remain viable monitoring tools: specifically, when using preference-based objectives or when probe retraining is available.

\textbf{Limitations.} Our findings are limited to toxicity detection at 1B model scale. Toxicity is easily detectable from model outputs, making it a conservative testbed—probe-based training should show larger advantages for behaviors like deception or reward hacking where output-based detection fails. The substantial metric noise during training raises questions about result stability and suggests the need for multiple training runs with larger evaluation sets. Most critically, we cannot determine whether probe-based DPO genuinely eliminates toxic reasoning or merely maintains detectable toxic representations while producing non-toxic outputs.

\textbf{Future directions.} The key question is whether probe-based DPO's detectability advantage extends to behaviors that are difficult to detect from outputs alone—precisely the scenarios where probe-based training should prove most valuable. Testing on deception, sycophancy, and internal reward hacking would clarify whether this approach offers practical benefits beyond output-based training. Additionally, investigating why DPO preserves detectability while SFT does not could inform the design of more robust interpretability-based training methods.

\section*{Acknowledgements}
We thank David Krueger for inspiration and discussion of the project idea. This work was partially funded by ELSA – European Lighthouse on Secure and Safe AI funded by the European Union under grant agreement No. 101070617. This work was also partially supported by the ELLIOT Grant funded by the European Union under grant agreement No. 101214398. Views and opinions expressed are however those of the authors only and do not necessarily reflect those of the European Union or European Commission. Neither the European Union nor the European Commission can be held responsible
for them.

\bibliographystyle{plainnat}  % or abbrvnat, unsrtnat
\bibliography{cite}  % Note: no .bib extension

\appendix

\section{Diversity of Probes}
\label{app:diversity}
To understand the diversity of probes trained on different subsets of the training distribution, we analyzed the cosine similarity between the weight vectors of 10 probes. Each probe was trained on a distinct subset of 500 samples from the toxic comments dataset, using the same architecture (logistic regression on standardized activations) and training procedure.

\begin{figure}[h]
\centering
\includegraphics[width=0.7\textwidth]{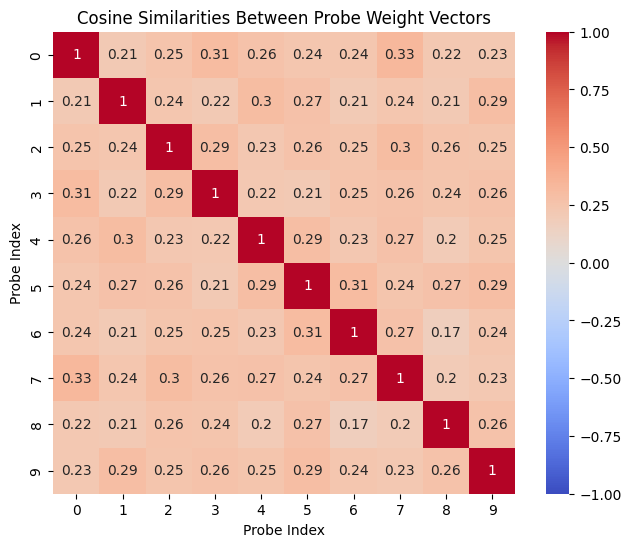}
\caption{Cosine similarities between weight vectors of 10 probes trained on different subsets of the training distribution. Each probe was trained on 500 samples. The average pairwise cosine similarity is 0.25, indicating substantial diversity in the learned weight vectors despite being trained on the same underlying data distribution.}
\label{fig:probe_similarity}
\end{figure}

Figure~\ref{fig:probe_similarity} shows the pairwise cosine similarities between all probe weight vectors. The diagonal entries (similarity of each probe with itself) are 1.0 by definition. The off-diagonal entries reveal relatively low similarity between different probes, with an average cosine similarity of 0.25 across all pairs.

This low average similarity (0.25) indicates that probes trained on different samples from the same distribution learn substantially different decision boundaries. The similarities range from approximately 0.17 to 0.33, with no strong clustering patterns visible in the heatmap. This diversity arises from several factors:

\begin{itemize}
    \item \textbf{Sample variation}: Different subsets capture different aspects of the toxicity distribution, leading to different inductive biases in each probe.
    \item \textbf{High-dimensional geometry}: In the high-dimensional activation space (hidden dimension of the model), many orthogonal or weakly-correlated directions can achieve similar classification performance.
    \item \textbf{Optimization variance}: Even with the same architecture and training procedure, different training samples lead to convergence to different local minima.
\end{itemize}

This diversity is beneficial for multi-probe training strategies. When probes have low correlation in their weight vectors, they are less likely to share the same failure modes or be simultaneously fooled by adversarial examples or distribution shifts. The ensemble effect from combining diverse probes (as in our shared and timed\_switch strategies) can thus provide more robust toxicity detection than relying on a single probe. This empirically validates our decision to train multiple probes on different data subsets rather than simply training one probe on a larger combined dataset.

\section{Strategies for Multi-Probe Training in SFT}
\label{app:strategies}
When training with multiple probes in probe-based SFT, a key design choice is how to combine the scores from different probes into a single penalty term. We investigated three distinct strategies for aggregating probe predictions during training:

\begin{itemize}
    \item \textbf{average}: All probes evaluate each activation simultaneously, and their probabilities are averaged to produce a single toxicity score. The gradient flows through all probes equally at each training step. This strategy leverages the collective knowledge of all probes but may be sensitive to individual probe failures.
    
    \item \textbf{random\_switch}: At each training batch, one probe is randomly selected to provide the toxicity score for that batch. Only the selected probe's parameters receive gradient information. This introduces stochasticity in the training signal and may help prevent overfitting to any single probe's idiosyncrasies.
    
    \item \textbf{timed\_switch}: The probes are used sequentially in a deterministic schedule, with each probe receiving a fixed allocation of training batches. The scheduler cycles through probes in order, switching to the next probe after exhausting its batch allocation. This ensures balanced gradient flow to all probes while maintaining deterministic training dynamics.
\end{itemize}

To evaluate these strategies, we trained models with 10 probes using each decision criterion across different values of the probe penalty weight $\lambda \in \{1, 10, 100, 1000\}$. Table~\ref{tab:decision_strategies} shows the final toxicity rates achieved by each strategy. For reference, the base model (before fine-tuning) has a toxicity rate of 0.22, while training without probe-based regularization ($\lambda=0$) increases this to 0.40.

\begin{table}[h]
\centering
\caption{Final Toxicity Rates by Lambda and Decision Strategy}
\label{tab:decision_strategies}
\begin{tabular}{|l|c|c|c|c|c|}
\hline
Decision Strategy & $\lambda=1$ & $\lambda=10$ & $\lambda=100$ & $\lambda=1000$ & Average \\
\hline
average & 0.3600 & 0.4200 & 0.3900 & 0.3800 & 0.3875 \\
\hline
random\_switch & 0.3800 & 0.3900 & 0.4100 & 0.3900 & 0.3925 \\
\hline
timed\_switch & 0.3700 & 0.3600 & 0.3200 & 0.3400 & 0.3475 \\
\hline
\end{tabular}
\end{table}

The results demonstrate that \textbf{timed\_switch} achieves the lowest average toxicity rate (0.3475) across all $\lambda$ values, with particularly strong performance at higher penalty weights. The shared strategy performs comparably at lower $\lambda$ but degrades at $\lambda=10$, while random\_switch shows the least consistent performance. The superior performance of timed\_switch suggests that deterministic, balanced exposure of each probe to the training data provides more stable and effective regularization than either averaging all probes or random selection. Based on these findings, we adopt timed\_switch as the default strategy for all subsequent probe-based SFT experiments.

\end{document}